\documentclass[runningheads]{llncs}
\pdfoutput=1
 
\usepackage{eccv}

\usepackage{eccvabbrv}

\usepackage{graphicx}
\usepackage{booktabs}
\usepackage{orcidlink} 
\usepackage{colortbl}
\usepackage{multirow}
\usepackage[export]{adjustbox}
\usepackage{subcaption}
\usepackage{bm}

\newcommand{\method}{MixDQ\xspace}

\makeatletter
\renewcommand\@fnsymbol[1]{%
    \ifcase#1\or
    *\or 
    \textdagger\or
    \textdaggerdbl\or
    \fi
}
\makeatother

\usepackage{amsmath}

\usepackage[accsupp]{axessibility} 
\usepackage[ruled,vlined]{algorithm2e}

\usepackage{hyperref}

\begin{document}

\title{MixDQ: Memory-Efficient Few-Step Text-to-Image Diffusion Models with Metric-Decoupled Mixed Precision Quantization} 

\titlerunning{MixDQ}


\author{Tianchen Zhao\inst{12}\thanks{Equal contribution} \and
Xuefei Ning\inst{1*}\thanks{Corresponding Authors: Yu Wang (yu-wang@mail.tsinghua.edu.cn), Xuefei Ning (foxdoraame@gmail.com)} \and
Tongcheng Fang\inst{12^*}
\and Enshu Liu\inst{1} \and \\
 Guyue Huang \inst{3} \and Zinan Lin \inst{4} \and Shengen Yan \inst{2} \and Guohao Dai \inst{25} \and Yu Wang\inst{1\text{\textdagger}}}


\authorrunning{Zhao et al.}

\institute{Tsinghua University \and Infinigence AI \and 
University of California Santa Barbara \and Microsoft 
\and Shanghai Jiaotong University}

\maketitle

\begin{abstract}

Few-step diffusion models, which enable high-quality text-to-image generation with only a few denoising steps, have substantially reduced inference time. However, considerable memory consumption (5-10GB) still poses limitations for practical deployment on mobile devices.
Post-Training Quantization (PTQ) proves to be an effective method for enhancing efficiency in both memory and operational complexity. However, when applied to few-step diffusion models, existing methods designed for multi-step diffusion face challenges in preserving both visual quality and text alignment. 
In this paper, we discover that the quantization is bottlenecked by highly sensitive layers. Consequently, we introduce a mixed-precision quantization method: \textbf{MixDQ}. 
Firstly, we identify some highly sensitive layers are caused by outliers in text embeddings, and design a specialized Begin-Of-Sentence (BOS)-aware quantization to address this issue.
Subsequently, we investigate the drawback of existing sensitivity metrics, and introduce metric-decoupled sensitivity analysis to accurately estimate sensitivity for both image quality and content. 
Finally, we develop an integer-programming-based method to obtain the optimal mixed-precision configuration.
In the challenging 1-step Stable Diffusion XL text-to-image task, current quantization methods fall short at W8A8. Remarkably, MixDQ achieves \textbf{W3.66A16} and \textbf{W4A8} quantization with negligible degradation in both visual quality and text alignment. Compared with FP16, it achieves \textbf{3-4$\times$} reduction in model size and memory costs, along with a \textbf{1.5$\times$} latency speedup.

\end{abstract}

\begin{figure}[t]
    \centering
    \includegraphics[width=1.0\textwidth]{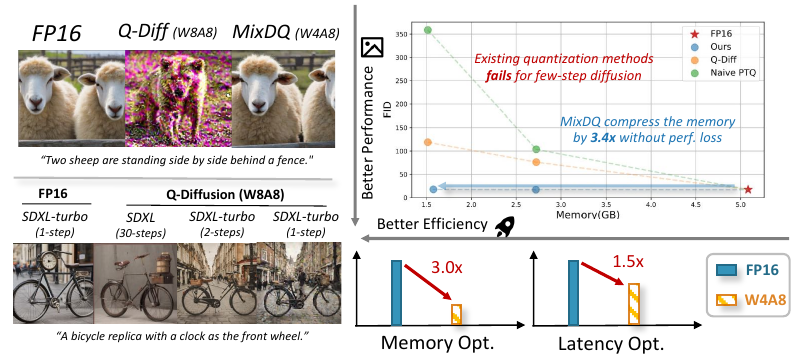}
    \caption{\textbf{The effectiveness of MixDQ.} Left: MixDQ preserves both image quality and image-text alignment. Right: The efficiency improvements of MixDQ. 
    }
    \label{fig:teaser}
\end{figure}

\section{Introduction}
\label{sec:intro}

Text-to-image diffusion models~\cite{stable-difffusion,DALL-E,sdxl} 
attract substantial research attention due to their capability to generate high-quality images from textual prompts. However, their demanding computational and memory requirements present challenges for real-time application and deployment on mobile devices~\cite{speed-is-all-you-need,mobile-diffusion}. Recent efforts on few-step diffusion models~\cite{cm,lcm-lora,sdxl-turbo} have significantly alleviated the computational burden, which require only 1-8 steps of iteration to generate high-fidelity images, as opposed to the 10-100 steps needed in previous approaches~\cite{ddpm,ddim}. However, the memory cost of diffusion models remains excessive~\cite{quest,personalization-quant}. For example, running a Stable Diffusion XL-turbo model~\cite{sdxl} in the 16-bit floating-point (FP16) format with a batch size of 1 consumes a peak memory of 9.7GB, which exceeds the capacity of many mobile devices and even some desktop GPUs (e.g., an RTX 4070 has 8GB GPU memory).

Model quantization~\cite{Jacob-quantization}, compressing high bit-width floating-point parameters and activations into lower bit-width integers, proves to be an effective strategy for reducing both memory and computational cost. Many prior researches~\cite{PTQD,efficientDM} explore quantization for diffusion models. 
However, we observe that \textbf{few-step diffusion models are more sensitive to quantization than multi-step ones}. As depicted in Fig.~\ref{fig:teaser}, the quantized SDXL 30-step model and SDXL-turbo 2-step model have notably less degradation compared with the SDXL-turbo 1-step model. Prior research Q-diffusion~\cite{q-diffusion} performs well on multi-step models but encounters challenges with W8A8 quantization for one-step SDXL-turbo\cite{sdxl-turbo}. The resulting image appears blurred and contains numerous artifacts. We conjecture this might be due to the absence of the iterative denoising process, which could compensate for the quantization error.

\begin{figure}[t]
    \centering
    \includegraphics[width=0.9\textwidth]{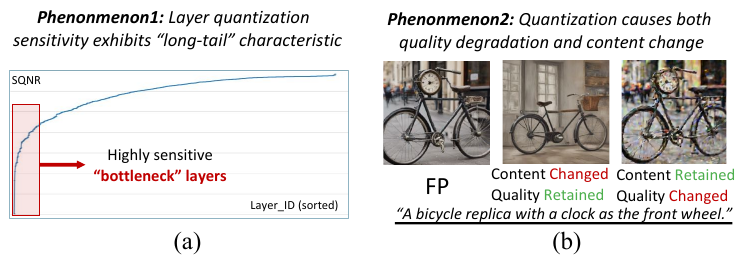}
    \caption{\textbf{Insightful findings for few-step text-to-image diffusion quantization.} Left: The layer-sensitivity distribution has a ``long-tail'' characteristic. Right: Quantization affects both image quality and content.
    }
    \label{fig:observation}
\end{figure}

Additionally, prior research primarily focuses on preserving the image quality for quantized models. However, as illustrated in \cref{fig:teaser} and Fig.~\ref{fig:observation}, quantization impacts not only image quality but also content. The altered content may lead to \textbf{degradation in image-text alignment}, which refers to how well the generated image aligns with the given text instruction. 
For instance, in \cref{fig:teaser}, the image generated by Q-diffusion depicts a polar bear, which contradicts the prompt mentioning ``goats''. Similarly, the middle image in \cref{fig:observation} loses the content ``clock''.

To effectively quantize the few-step model while preserving the quality and alignment in the meantime, we investigate the reasons behind the failure of existing quantization methods. We discover two insightful phenomena, as presented in \cref{fig:observation}. Firstly, we measure each layer's quantization sensitivity for the SDXL-turbo model with the signal-to-quantization-noise ratio (SQNR) following prior research~\cite{sqnr-ptq}. As shown in \cref{fig:observation} (a), the distribution exhibits a ``long-tail'' characteristic. Consequently, using uniform quantization settings for all layers, as in previous research, would result in the quantization being ``bottlenecked'' by some highly sensitive layers.
We look into these sensitive layers and propose: (1) designing a specialized quantization method to ``protect'' the outliers in certain layers, significantly reducing their quantization error; and (2) adopting mixed-precision quantization to ``protect'' the sensitive layers with a higher bit-width, thus achieving higher quality.
Secondly, as illustrated in \cref{fig:observation} (b), quantization affects both the image content and quality. We identify that the entanglement of these two factors in evaluation will cause quantization methods to fail in preserving generation quality. Therefore, we propose ``decoupling'' the effects of quality and content in the sensitivity evaluation. This will be elaborated in \cref{sec:metric-decouple}.

In light of above, we introduce a mixed-precision quantization method: \textbf{MixDQ}.
Firstly, we analyze the characteristics of highly sensitive layers and discover that many of them are associated with the quantization of text embeddings. We further investigate the feature distribution of text embeddings and design a specialized \textbf{Begin-of-sentence (BOS)-aware quantization} (Sec.~\ref{sec:bos}) to address the outlier values.
Secondly, we identify the shortcomings of existing quantization sensitivity evaluation 
and design an improved \textbf{Metric-decoupled sensitivity analysis} (Sec.~\ref{sec:metric-decouple}) based on the idea of decoupling the impact of quantization on image content and quality.
Finally, we design an \textbf{integer-programming-based mixed precision allocation} (Sec.~\ref{sec:integer-programming}) to acquire optimal bit-width configuration based on the improved sensitivity analysis. 

We summarize the contributions of this paper as follows:

\begin{enumerate}
    \item We highlight the challenge of quantizing the few-step diffusion model, compared to quantizing the multi-step diffusion model. Additionally, we emphasize the often-overlooked necessity of preserving alignment when compressing the text-to-image generative models. 
    \item Based on the careful investigation of the data distribution and sensitivity of each layer, we design MixDQ, a mixed precision quantization method with improved sensitivity evaluation and quantization techniques.
    \item We evaluate MixDQ in two settings: weight-only quantization and normal weight and activation quantization. MixDQ can achieve W4A16 and W5A8 quantization with a negligible 0.1 FID increase. Compared with the FP16 model, MixDQ can achieve W3.66A16 and W4A8 quantization within a 0.5 FID increase, resulting in a 3$\times$ memory cost reduction and a 1.5$\times$ latency speedup on Nvidia GPUs.
\end{enumerate}

The techniques in MixDQ could benefit future research and applications of compression methods on other generative models and tasks. Firstly, BOS-aware quantization addresses the issue of outlier values in text embeddings, which is an inherent problem for transformer-based models (also recognized as the ``attention sink''~\cite{attention-sink} in language models). Secondly, we believe the explicit and decoupled consideration of various metrics is important whenever compressing models on broader visual generation tasks. On one hand, from an application perspective, multiple metrics should be considered, especially for generative tasks. On the other hand, for better compression results, the explicit and decoupled consideration of multiple metrics can avoid the failure of compression methods. 
We will release all the code and models.

\section{Related Work}
\label{sec:related_works}


\noindent
\textbf{Diffusion Models.}
~\cite{2015diffusion,ddpm,sde} could generate high-quality images through an iterative denoising process. However, the excessive cost of repeated denoising iterations calls for improvements in the sampling efficiency, namely, reducing the timesteps. Some research~\cite{ddim,dpm-solver,unipc} focuses on designing improved numerical solvers, and another line of research explores utilizing distillation~\cite{pd,on_distill,lcm-lora,sdxl-turbo} to condense the sampling trajectory into fewer steps or even a single step. These few-step diffusion models significantly reduce the computational cost, however, as discussed in Sec.~\ref{sec:intro}, \textbf{they face additional challenges for quantization}.  \\


\noindent
\textbf{Network Quantization.}
Prior research, such as PTQD~\cite{PTQD} and Q-DM~\cite{q-dm} explores utilizing quantization techniques on diffusion models. Q-Diffusion~\cite{q-diffusion} extends this application to large-scale stable diffusion models. Other research~\cite{pcr,temporal-quant-1,temporal-quant-2} continues to improve post-training quantization techniques on aspects like calibration and temporal adjustment. 
However, the majority of existing diffusion quantization methods \textbf{employs uniform bit-with for layers with diverse sensitivity}. Yang et. al.~\cite{sqnr-ptq} suggest preserving sensitive layers as high-precision based on SQNR metric. However, we observe that the SQNR metric tends to prioritize content change, potentially leading to quality degradation (as discussed in Sec.~\ref{sec:metric-decouple}). Inspired by the prior success of mixed precision~\cite{haq,hawq}, we design a mixed precision quantization method to address the imbalanced sensitivity. \\


\noindent
\textbf{Evaluation Metrics.}
The currently widely used metrics for evaluating text-to-image generation can be summarized into two major aspects: image fidelity (quality) and image-text alignment. To estimate image fidelity, metrics such as FID~\cite{FID} and IS~\cite{IS} are commonly used. These metrics measure the feature space distance between generated images and reference images. For image-text alignment, CLIP Score~\cite{CLIPScore} is often utilized to calculate the similarity of 
the image and the text embeddings.
 To align the statistical scores and human preference, ImageReward~\cite{ImageReward} and HPS~\cite{HPS} collect user preference and train the model to predict generated image scoring, considering both fidelity and alignment. While previous methods solely discuss preserving the image quality, we investigate quantization's effect on \textbf{both image fidelity and alignment}. 

\begin{figure}[t]
    \centering
    \includegraphics[width=0.95\textwidth]{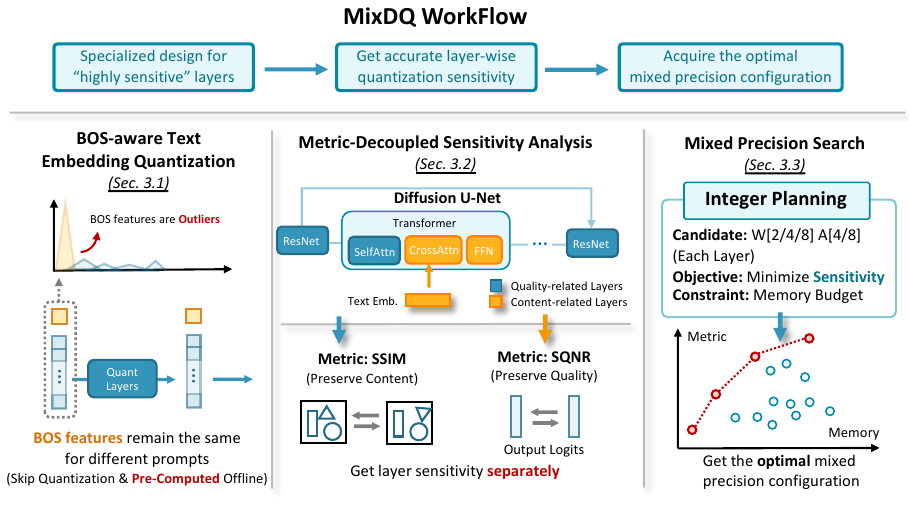}
    \caption{\textbf{Framework of the proposed mixed-precision quantization method: MixDQ.} It consists of three key components, the BOS-aware quantization addresses the highly sensitive text embedding, the metric-decoupled scheme improves sensitivity analysis, and the integer programming acquires the optimal bit-width allocation.}
    \label{fig:method}
\end{figure}

\section{Methods}
\label{sec:method}


Fig.~\ref{fig:method} presents our mixed precision quantization method \textbf{MixDQ} consisting of three consecutive steps. Firstly, we identify the highly sensitive layers and examine their distinctive properties. Based on these properties, we devise specialized quantization techniques tailored for them (Sec.~\ref{sec:bos}). 
Secondly, in \cref{sec:metric-decouple}, we introduce a metric-decoupled analysis that measures the quantization's effects on image content and quality separately.
Finally, based on the sensitivity, we employ integer programming to determine the optimal mixed precision configuration under a given budget (Sec.~\ref{sec:integer-programming}). 

\begin{figure}[t]
    \centering
    \includegraphics[width=0.8\textwidth]{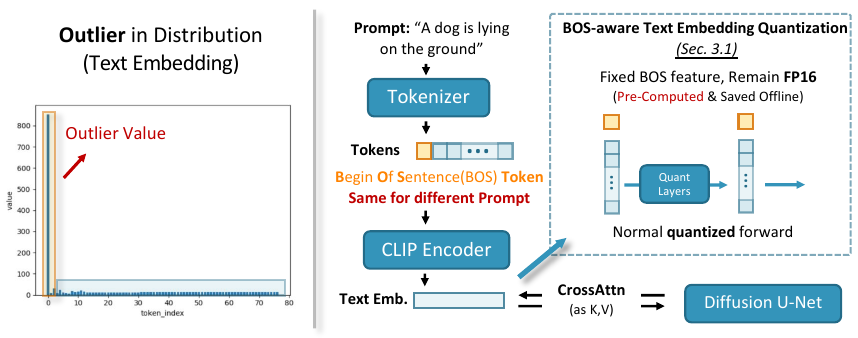}
    \caption{\textbf{Illustration of BOS-aware Quantization.} Left: the first token has a significantly larger value than the others. Right: Since BOS token features remain the same for different prompts, we skip quantizing them and pre-compute them offline.}
    \label{fig:bos}
\end{figure}

\subsection{BOS-aware Text Embedding Quantization}
\label{sec:bos}


Fig.~\ref{fig:observation} shows that the diffusion model quantization is ``bottlenecked'' by some highly sensitive layers.
Upon going through the highly sensitive layers depicted in the ``long-tail'', we find that a substantial portion of them corresponds to the ``to\_k'' and ``to\_v'' linear layers in the cross-attention.
All of these layers take the text embedding, which is the output of the CLIP encoder~\cite{CLIP}, as their input. Fig.~\ref{fig:bos} (Left) shows the maximum magnitude of the text embedding of each token (average across 8 sentences). We observe that the 1st token has a significantly larger magnitude (823.5) compared to the rest of the tokens (10-15).
Quantizing this tensor with such an outlier value to 8-bit would lead to the majority of values being close to 0, resulting in the loss of crucial textual information. This observation could potentially explain why current quantization methods struggle to maintain text-image alignment in Fig.~\ref{fig:teaser} (Left).

As shown in Fig.~\ref{fig:bos} (Right), the first outlier token in the CLIP output corresponds to 
the ``Begin Of Sentence (BOS)'' in the tokenizer output. Actually, \emph{the feature of this BOS token remains the same across different prompts}.
Therefore, we can skip the quantization and computation for the ``to\_k/v'' layers for this token. Concretely, we pre-compute the floating-point output feature of the ``to\_k/v'' layers for the BOS token and then concatenate it with the dequantized features of other tokens. In this way, the quantization error of the CLIP embedding is significantly reduced as the outlier is removed.
Additionally, only 640-1280 (the channel number of BOS features) elements need to be stored for each ``to\_k/v'' layer, introducing only negligible overhead.

\begin{figure}[t]
    \centering
    \includegraphics[width=1.0\textwidth]{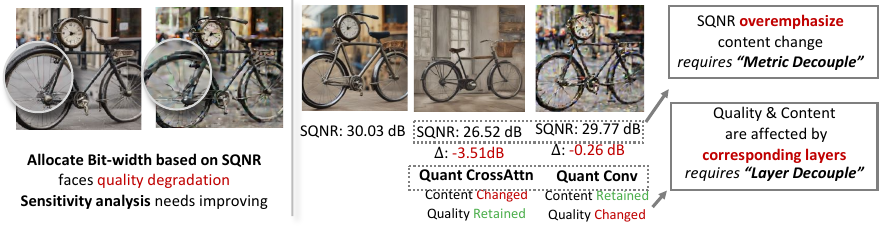}
    \caption{\textbf{Decoupling metrics and layers to separate the influence on image quality and content.} Left: Existing SQNR-based sensitivity analysis needs improving. Right: SQNR's problem of overemphasizing content change.
    }
    \label{fig:decouple}
\end{figure}

\subsection{Metric-Decoupled Sensitivity Analysis}
\label{sec:metric-decouple}


\textbf{Preliminary Experiment and Analysis.} We start by designing preliminary experiments.
As discussed in \cref{sec:intro}, we measure the sensitivity with SQNR. It is computed as the ratio of the L2 norm of feature values to the quantization noise. The quantization noise is estimated by evaluating the impact on the network's output logits with each layer quantized.
We observe the ``long-tailed'' characteristics of sensitivity distribution in \cref{fig:observation} (a). An intuitive solution is to allocate higher bit-widths for highly sensitive layers and lower bit-widths for less sensitive ones. We implement a straightforward bit-width allocation method based on this principle to achieve an average of W8A8. However, as seen in \cref{fig:decouple} (Left), the generated image faces severe quality degradation, suggesting that the current sensitivity evaluation relying solely on SQNR may require refinement. 

We look into this issue and observe the following phenomena in \cref{fig:decouple} (Right): (1) Both changes in image content and quality impact the SQNR. In particular,
compared with the 30.03dB image, the SQNR decline of an image with changed contents but high quality (-3.51dB) is much more significant than that of an image with similar contents but unacceptable quality (-0.26dB).
This means that \emph{when evaluating images with varying content and quality, SQNR tends to underestimate the performance degradation resulting from decreased image quality.}
(2) \emph{Image quality and content are mainly affected by different layer types.} Specifically, quantizing the cross-attention layers causes content changes, but the quality remains good. This aligns with the intuition that text supervision predominantly interacts with cross-attention layers, which control the image content. Conversely, 
when quantizing convolutions, the quality degrades and the content is preserved. 
Further details of the correlation between layer types and their impact on content or quality are provided in Appendix \ref{sec:Justification of Metric Decouple}.

As a result, during the bit-width allocation process, since SQNR tends to overemphasize content changes, the content-related layers are retained at very high bit-widths. Consequently, the bit-width for quality-related layers is decreased. This \emph{``unfair competition'' between quality- and content-related layers} explains the severe degradation of the image quality observed in \cref{fig:decouple} (Left) when the bit-width allocation is solely based on SQNR.

\noindent\textbf{Method Design.} Inspired by these findings, we improve the sensitivity analysis by decoupling metrics and layers to separate the quantization's effect on quality and content.  Therefore, we separate the layers into two groups: the content-related layers, i.e., the cross-attention layers and feed forward networks (FFNs), and the quality-related layers, i.e., self-attention layers and convolutions. Then, we conduct sensitivity analysis for each group separately with distinct metrics. 

When performing the sensitivity analysis for the content-related layers,
we adopt the Structural Similarity Index Measure (SSIM)~\cite{SSIM} metric to assess image content change. Given a generated image $x$ and reference image $y$, The SSIM metric combines the three components, the luminance ($l$), the contrast ($c$) and the structure ($s$):
\begin{equation}
    \resizebox{0.7\linewidth}{!}{
     $\begin{aligned}
     l(x,y) = \frac{2\mu_x\mu_y}{\mu_x^2+\mu_y^2}, 
     & c(x,y) = \frac{2\sigma_x\sigma_y}{\sigma_x^2+\sigma_y^2},
     s(x,y) = \frac{\sigma_{xy}}{\sigma_x\sigma_y}, \\
    \mbox{SSIM}(x,y) &= l(x,y)^{\alpha} \cdot c(x,y)^{\beta} \cdot s(x,y)^{\gamma},   
    \end{aligned}$
    }
\label{equ:ssim}
\end{equation}
where $\mu_, \sigma$ stand for the mean and variance of pixel values; and $\alpha,\beta,\gamma$ are weights that control the combination of three components. We set $\alpha=\beta=\gamma=1$, and use the full image size as the window size for SSIM. 
Unlike Mean Squared Error (MSE) and Peak Signal-to-Noise Ratio (PSNR) that measure absolute errors, SSIM is adept at perceiving changes in structural information, making it well-suited for measuring content change.

When conducting sensitivity analysis for the quality-related layers, we utilize the SQNR metric. Given that quantizing only the quality-related layers does not significantly alter the image content, SQNR serves as a suitable metric for assessing the degree of image quality degradation in this scenario.



\subsection{Integer Programming Bit-width Allocation}
\label{sec:integer-programming}

Having obtained the layer sensitivity, we can determine the mixed precision configuration by assigning higher bit-widths to more sensitive layers. Enlightened by prior research~\cite{llm_mq}, we formulate the bit-width allocation into an integer programming problem: Given the resource budget $\mathcal{B}$, and candidate bit-widths $b \in \{2,4,8\}$, we aim to determine the bit-width choices $\bm{c}$ (a one-hot indicator for each layer) that maximizes the sum of sensitivity $\mathcal{S}=\sum_{i=1}^{N} \sum_{b=2,4,8} c_{i,b} \cdot \mathcal{S}_{i,b}$: 
\begin{equation}
    \resizebox{0.6\linewidth}{!}{
    $\begin{aligned}
    &\underset{c_{i,b}}{\text{argmax}} \quad \sum_{i=1}^{N} \sum_{b=2,4,8} c_{i,b} \cdot \mathcal{S}_{i,b} \\
    &\text{s.t.} \quad \sum_{b=2,4,8} c_{i,b} = 1, \quad \sum_{i=1}^{N} \sum_{b=2,4,8} c_{i,b} \cdot \mathcal{M}_{i,b} \leq \mathcal{B}, \\
    &\quad c_{i,b} \in \{0,1\}, \quad \forall i \in \{1,\cdots,N\}, \forall b \in \{2,4,8\},
    \end{aligned}$
    }
\label{equ:integer-planning}
\end{equation}
where $N$ is the number of layers in the model; $c_{i,b}=1$  indicates that the $i$-th layer will be quantized to $b$-bit,
and $\mathcal{S}_{i,b}$ is the corresponding sensitivity score (higher SQNR or SSIM is better).
$\mathcal{M}_{i,b}$ denotes the resource cost of the $i$-th layer when it is quantized to $b$-bit. 

We conduct the above integer programming separately for activation quantization and weight quantization, as well as for each of the two layer groups, i.e., the content-related cross-attention and FFN layers, and the quality-related self-attention and convolution layers. 


\section{Experiments}
\label{sec:exp}

\subsection{Experimental Settings}
\label{sec:exp:exp_setting}

\noindent
\textbf{Quantization Scheme.}
We adopt the simplest and easily deployable asymmetric min-max quantization scheme similar to~\cite{practical-mixed-precision, quant-white-paper}. 
The quantization parameters (i.e., scaling factor, zero point) are shared within each tensor for the activation or within each output channel for the weight.
The shortcut-splitting quantization technique~\cite{q-diffusion} is applied to all methods in Tab.~\ref{tab:main}. 
We randomly sampled 1024 prompts from COCO~\cite{COCO} as our calibration dataset.

\noindent
\textbf{Mixed Precision Allocation.}
We do not quantize all nonlinear activation and normalization layers, and quantize all the linear and convolution layers.
In our metric-decoupled sensitivity analysis, we iterate through all layers one by one. For each layer, our method quantizes only that layer and preserves the remaining layers as FP. Then, we measure two types of sensitivity metrics for this layer: SQNR and SSIM. The sensitivity score is averaged over 32 prompts. We quantize each layer to three bit-width choices: 2, 4, 8, and get their corresponding sensitivity scores.
Then, we set a budget for the average bit-width of all elements (weights or activation) and use the layer-wise sensitivity scores to set up the integer programming. We use the OR-tools~\cite{or-tools} library for integer programming. The efficiency of this implementation allows the bit-width allocation in seconds.
For both layer groups (content-related and quality-related), we sweep 20 choices of the average bit-width budget, ranging uniformly from 3 to 8. Then, we keep the Pareto frontier of the 20$\times$20=400 bit-width configurations (Fig.~\ref{fig:pareto}). 


\noindent
\textbf{Hardware Profiling.}
We measure the latency and memory usage of MixDQ on the Nvidia RTX 4080 GPU using CUDA 12.1. All profiling is conducted with a batch size of 1. Specifically, we measure memory usage for all models using the PyTorch Memory Management APIs~\cite{pytorchmemorymanagement}. The inference latency of the FP16 models is profiled with NVIDIA Nsight tools~\cite{nsightsystem}. For quantized models, we measure the latency of the quantized layers, the quantization operation, and the unquantized layers separately, and add them up to calculate the overall latency.
We profile the FP and quantized models using stable-fast~\cite{stablefast}, a toolkit that provides state-of-the-art inference speed for diffuser models. We use the kernels from the Cutlass~\cite{cutlass} library to implement the quantized layers and develop a quantization GPU kernel to reduce the quantization overhead.

%

\begin{table*}[t]
\centering
\caption{\textbf{Performance and efficiency comparison of MixDQ and other quantization methods on full COCO annotations.} The ``CLIP'' and ``IR'' denotes CLIP Score and ImageReward metric. The ``Storage Opt.'' and ``Compute Opt.'' denote equivalent savings of model size and computational complexity (measured in Bit Operations as in~\cite{practical-mixed-precision}). The bit-width ``16'' represents FP16 without quantization. 
The ``weight only'' setting represents the rows with activation bit-width of FP16, and the rest are the ``normal'' weight-activation quantization.}
\resizebox{0.9\linewidth}{!}{
\label{tab:main}
\begin{tabular}{cccccccc}
\toprule[1pt]
\multirow{2}{*}{\textbf{Model}} & \multirow{2}{*}{\textbf{Method}} & \textbf{Bit-width} & \textit{Storage} & \textit{Compute} & \multirow{2}{*}{\textbf{FID($\downarrow$)}} & \multirow{2}{*}{\textbf{CLIP($\uparrow$)}} & \multirow{2}{*}{\textbf{IR($\uparrow$)}} \\
 &  & (W/A) & \textit{Opt.}  & \textit{Opt.} & & & \\
\midrule \midrule
\multirow{13}{*}{\begin{tabular}[c]{@{}c@{}}\textbf{SDXL-turbo}\\ (1 step)\end{tabular}} & FP & 16/16 & - & - & 17.15 & 0.2722 & 0.8631 \\
 \cmidrule(lr){2-8}
 & \multirow{4}{*}{Naive PTQ} & 8/16 & 2$\times$ & 1$\times$ & 16.89 & 0.2740 & 0.8550 \\
 &  & 4/16 & 4$\times$ & 1$\times$ & 301.49 & 0.1581 & -2.2526 \\
 &  & 8/8 & 2$\times$ & 4$\times$ & 103.96 & 0.1478 & -1.7446 \\
 &  & 4/8 & 4$\times$ & 8$\times$ & 358.894 & 0.1242 & -2.2815 \\
\cmidrule(lr){2-8}
 & \multirow{4}{*}{Q-Diffusion} & 8/16 & 2$\times$ & 1$\times$ & 16.97 & 0.2735 & 0.8588 \\
 &  & 4/16 & 4$\times$ & 1$\times$ & 22.58 & 0.2685 & 0.6847 \\
 &  & 8/8 & 2$\times$ & 4$\times$ & 76.18 & 0.1772 & -1.3112 \\
 &  & 4/8 & 4$\times$ & 8$\times$ & 118.93 & 0.1662 & -1.6353 \\
 \cmidrule(lr){2-8}
 & \multirow{5}{*}{MixDQ(Ours)} & 4/16 & 4$\times$ & 1$\times$ & 17.23 & 0.2693 & 0.8254 \\
 &  & 3.66/16 & 4.4$\times$ & 1$\times$ & 17.40 & 0.2682 & 0.7528 \\
 &  & 8/8 & 2$\times$ & 4$\times$ & 17.03 & 0.2703 & 0.8415 \\
 &  & 5/8 & 3.2$\times$ & 8$\times$ & 17.23 & 0.2697 & 0.8307 \\
 &  & 4/8 & 4$\times$ & 8$\times$ & 17.68 & 0.2698 & 0.7822 \\
\midrule
\multirow{7}{*}{\begin{tabular}[c]{@{}c@{}}\textbf{LCM-lora}\\ (4 steps)\end{tabular}} & FP & 16/16 & - & - & 25.56 & 0.2570 & 0.2122 \\
\cmidrule(lr){2-8}
 & \multirow{2}{*}{Naive PTQ} & 8/8 & 2$\times$ & 4$\times$ & 23.36 & 0.2548 & 0.0517 \\
 &  & 4/8 & 4$\times$ & 8$\times$ & 87.36 & 0.2055 & -1.6160 \\
\cmidrule(lr){2-8}
 & \multirow{2}{*}{Q-Diffusion} & 8/8 & 2$\times$ & 4$\times$ & 23.92 & 0.2561 & 0.1875 \\
 &  & 4/8 & 4$\times$ & 8$\times$ & 57.73 & 0.2280 & -1.1863 \\
\cmidrule(lr){2-8}
 & \multirow{2}{*}{MixDQ(Ours)} & 8/8 & 2$\times$ & 4$\times$ & 22.54 & 0.2552 & 0.1573 \\
 &  & 4/8 & 4$\times$ & 8$\times$ & 33.48 & 0.2403 & -0.6732 \\
\bottomrule[1pt]
\end{tabular}}
\vspace{-10pt}
\end{table*}

\begin{table*}[t]
\centering
\caption{\textbf{Comparison with other quantization techniques for W8A8 on COCO annotations subset.} The ``CLIP'' and ``IR'' denotes CLIP Score and ImageReward metric. The metrics are evaluated with the first 1024 prompts in COCO annotations. The ``MP-only'' represents adopting the mixed precision only.}
\resizebox{0.85\linewidth}{!}{
\label{tab:comparison_quant}
\begin{tabular}{ccccc}
\toprule[1pt]
\textbf{Model} & \textbf{Method} & \textbf{FID($\downarrow$)} & \textbf{CLIP($\uparrow$)} & \textbf{IR($\uparrow$)} \\
\midrule \midrule
\multirow{8}{*}{\begin{tabular}[c]{@{}c@{}}\textbf{SDXL-turbo}\\ (1 step)\end{tabular}} & FP16 & 84.51 & 0.26 & 0.84 \\
 \cmidrule(lr){2-5}
 & Naive PTQ & 165.92 (\textcolor{BrickRed}{+81.4}) & 0.15 (\textcolor{BrickRed}{-0.11}) & -1.72 (\textcolor{BrickRed}{-2.56}) \\
 & PTQD~\cite{PTQD} & 340.74 (\textcolor{BrickRed}{+256.2}) & 0.12 (\textcolor{BrickRed}{-0.14}) & -2.28 (\textcolor{BrickRed}{-3.12}) \\
 & Q-Diffusion~\cite{q-diffusion} & 149.15 (\textcolor{BrickRed}{+64.6}) & 0.16 (\textcolor{BrickRed}{-0.10}) & -1.69 (\textcolor{BrickRed}{-2.53}) \\
 & EDA-DM~\cite{eda-dm} & 137.98 (\textcolor{BrickRed}{+53.5}) & 0.16 (\textcolor{BrickRed}{-0.10}) & -1.71 (\textcolor{BrickRed}{-2.55}) \\
 & MP-only~\cite{sqnr-ptq} & 114.70 (\textcolor{BrickRed}{+30.2}) & 0.15 (\textcolor{BrickRed}{-0.11}) & -0.61 (\textcolor{BrickRed}{-1.45}) \\
 & Non-Uniform (FP8)~\cite{fp8} & 101.73 (\textcolor{BrickRed}{+17.2}) & 0.24 (\textcolor{BrickRed}{-0.01}) & 0.16 (\textcolor{BrickRed}{-1.45}) \\
 \cmidrule(lr){2-5}
 & MixDQ & 83.39 (\textcolor{ForestGreen}{-1.12}) & 0.27 (\textcolor{ForestGreen}{+0.01}) & 0.84 (\textcolor{ForestGreen}{+0.00}) \\
\bottomrule[1pt]
\end{tabular}}
\end{table*}

\subsection{Performance and Efficiency Comparison}
\label{sec:exp:exp_performance}
\begin{figure}[t]
    \centering
    \includegraphics[width=0.9\textwidth]{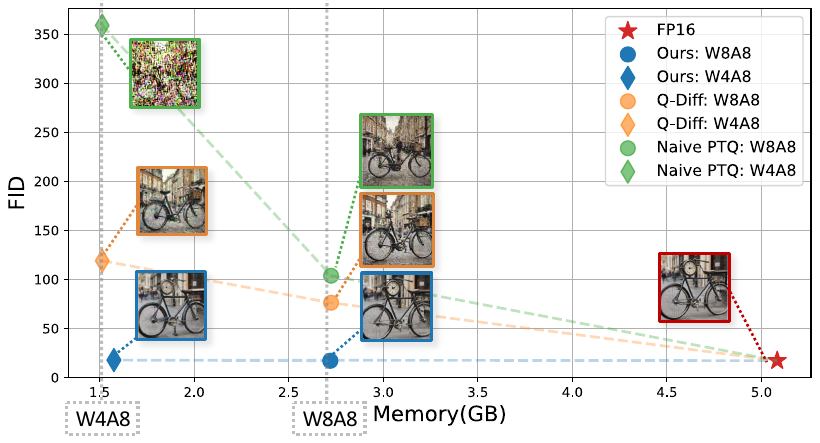}
    \caption{\textbf{The FID with respect to memory cost of MixDQ and baseline quantization methods, with corresponding generated images.} MixDQ achieves lossless quantization, whereas baseline methods fail to generate readable images. 
    }
    \vspace{-10pt}
    \label{fig:main-visual}
\end{figure}

We conduct experiments on widely-used few-step diffusion models, SDXL-turbo, and LCM-Lora, for text-to-image generation tasks using COCO2014 at the resolution of 512$\times$512. We adopt three different metrics: FID for fidelity, CLIP Score for image-text alignment, and ImageReward for human preference. The metrics are calculated on all 40,504 prompts. For the baseline methods, ``naive PTQ'' and ``Q-diffusion,'' we use a uniform bit-width for all layers. For MixDQ, we calculated the average bit-width weighted by each layer's parameter size. Following \cite{practical-mixed-precision}, we measure the theoretical computational savings (``Compute Opt.'') in Bit Operations (BOPs). The ``Storage Opt.'' represents the model size reduction.
The comparison of the performance and resource consumption of MixDQ quantization is presented in Tab.~\ref{tab:main} and Fig.~\ref{fig:main-visual}.

We experiment with two quantization settings: the ``weight-only'' scheme which seeks larger compression rate of model size, and the ``normal'' scheme focuses on both storage saving and latency speedup. During activation quantization, we observed that certain layers retain sensitivity and cannot be quantized to 8 bits without sacrificing performance. To address this, we choose to retain 1\% of the most sensitive layers based on metric-decoupled sensitivity. \cref{fig:hardware} shows that it introduces minimal overhead but ensures the preservation of performance.

As evident from Fig.~\ref{fig:main-visual} and Tab.\ref{tab:main}, for the SDXL-turbo model, the baseline quantization methods can only maintain image quality with W8A16. The naive PTQ's FID increases drastically from around 17.15 to 103.96 and 301.49 for W8A8 and W4A16, respectively. Q-diffusion, with the assistance of Adaround~\cite{adaround}, manages to preserve performance for W4A16 but still fails at W8A8, generating images with ``oil-painting''-like quality degradation. For W4A8, both PTQ and Q-diffusion produce images that are hardly readable, exhibiting FID values exceeding 100, negative ImageReward, and CLIP Score below 0.2. 
In contrast, MixDQ generates images that are nearly identical to FP16 images, maintaining both the image content and quality. 
Even with W4A8 quantization, MixDQ incurs a 0.5 FID increase and a 2.5e-3 CLIP Score drop. As for LCM-Lora, the baseline methods encounter fewer failures since they involve 4 iteration steps. Nevertheless, MixDQ consistently outperforms them for all metrics.

We further compare MixDQ's performance with other quantization techniques for W8A8 in Tab.~\ref{tab:comparison_quant}. 
For PTQD \cite{PTQD}, we conduct linear regression to determine $K$. However, we observe a relatively lower linear correlation (0.59) compared to the original paper. In the case of 1-step generation, the sampling process is deterministic, the "uncorrelated noise correction" inapplicable, and only the "correlated noise correction" is adopted. Despite being effective for multi-step quantization, we observe that PTQ causes notable quality degradation for the challenging 1-step quantization.
EDA-DM \cite{eda-dm} improves upon the quantization parameter tuning in Q-Diffusion \cite{q-diffusion} with layer-wise reconstruction. However, only a marginal improvement is witnessed. The term "MP-only" refers to "mixed precision only" with the SQNR-based method \cite{sqnr-ptq}, where due to suboptimal quantization sensitivity analysis, a significant performance degradation is still evident.
The "Non-uniform" refers to the use of the FP8 (Floating-point 8-bit) format \cite{fp8}, which introduces exponential bits to handle a larger dynamic range of data distribution. While this notably improves performance, it still falls short of the FP16 baseline. In contrast, MixDQ achieves W8A8 quantization without any performance loss.

\subsection{Hardware Resource Savings}
\label{sec:exp:hardware}
\begin{figure}[t]
    \centering
    \begin{minipage}{0.33\textwidth}
        \centering
        \resizebox{1.0\linewidth}{!}{
        \begin{tabular}{cccc}
            \toprule[1pt]
            Bit-width & Memory & Latency & \multirow{2}{*}{FID ($\downarrow$)}\\
            (W/A) & Opt. & Opt. \\
            \midrule \midrule
            16/16 & - & - & 17.15\\
            4/16 & 3.03$\times$ & - & 17.23\\
            8/8 & 1.87$\times$ & 1.45$\times$ & 17.03\\
            4/8 & 3.03$\times$ & 1.45$\times$ & 17.68\\
            \bottomrule[1pt]
        \end{tabular}
        }
    \end{minipage}
    \hfill
    \begin{minipage}{0.66\textwidth}
        \centering
        \begin{subfigure}[b]{0.5\textwidth}
            \centering
            \includegraphics[width=\textwidth]{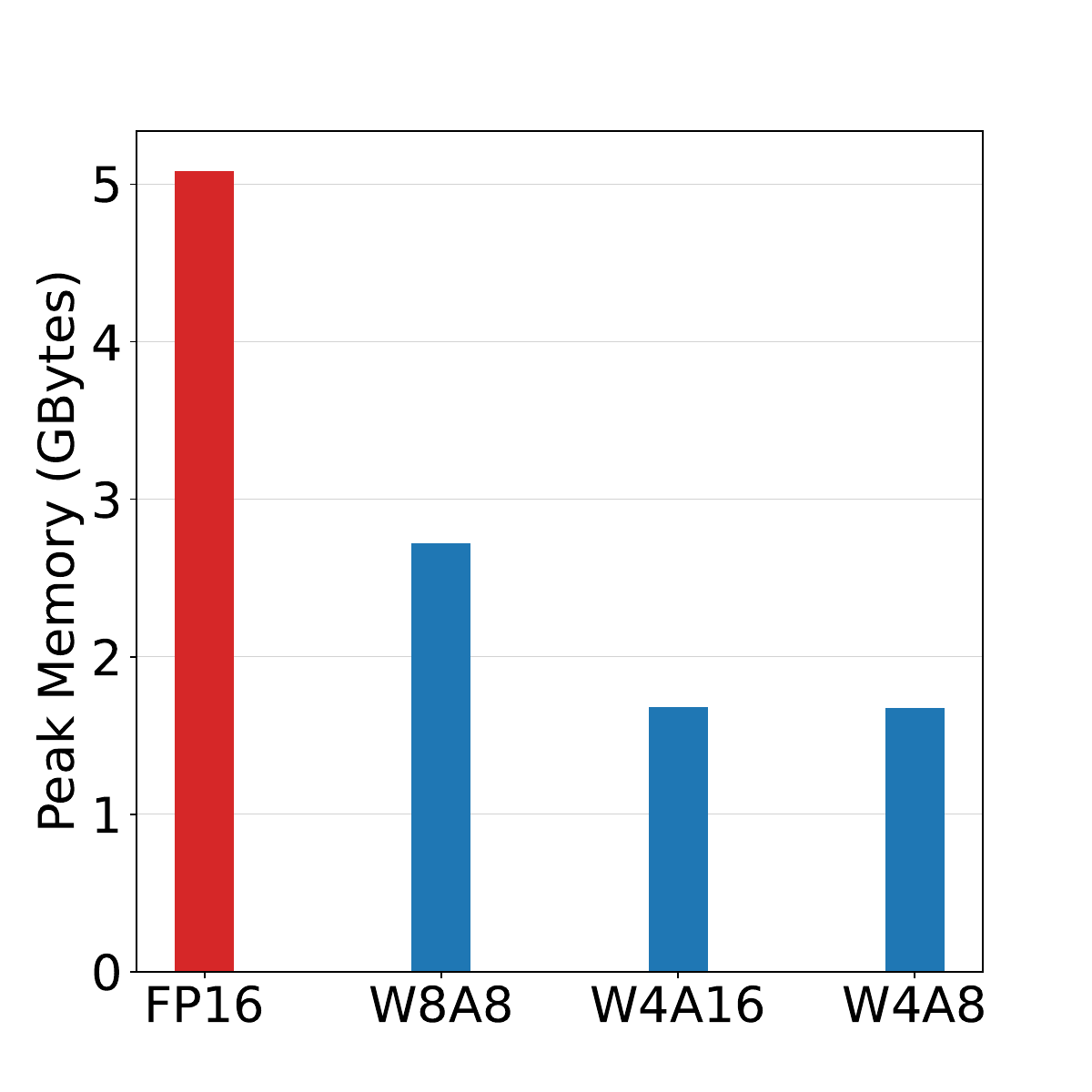}
            \caption{}
            \label{fig:hardware:a}
        \end{subfigure}%
        \begin{subfigure}[b]{0.5\textwidth}
            \centering
            \includegraphics[width=\textwidth]{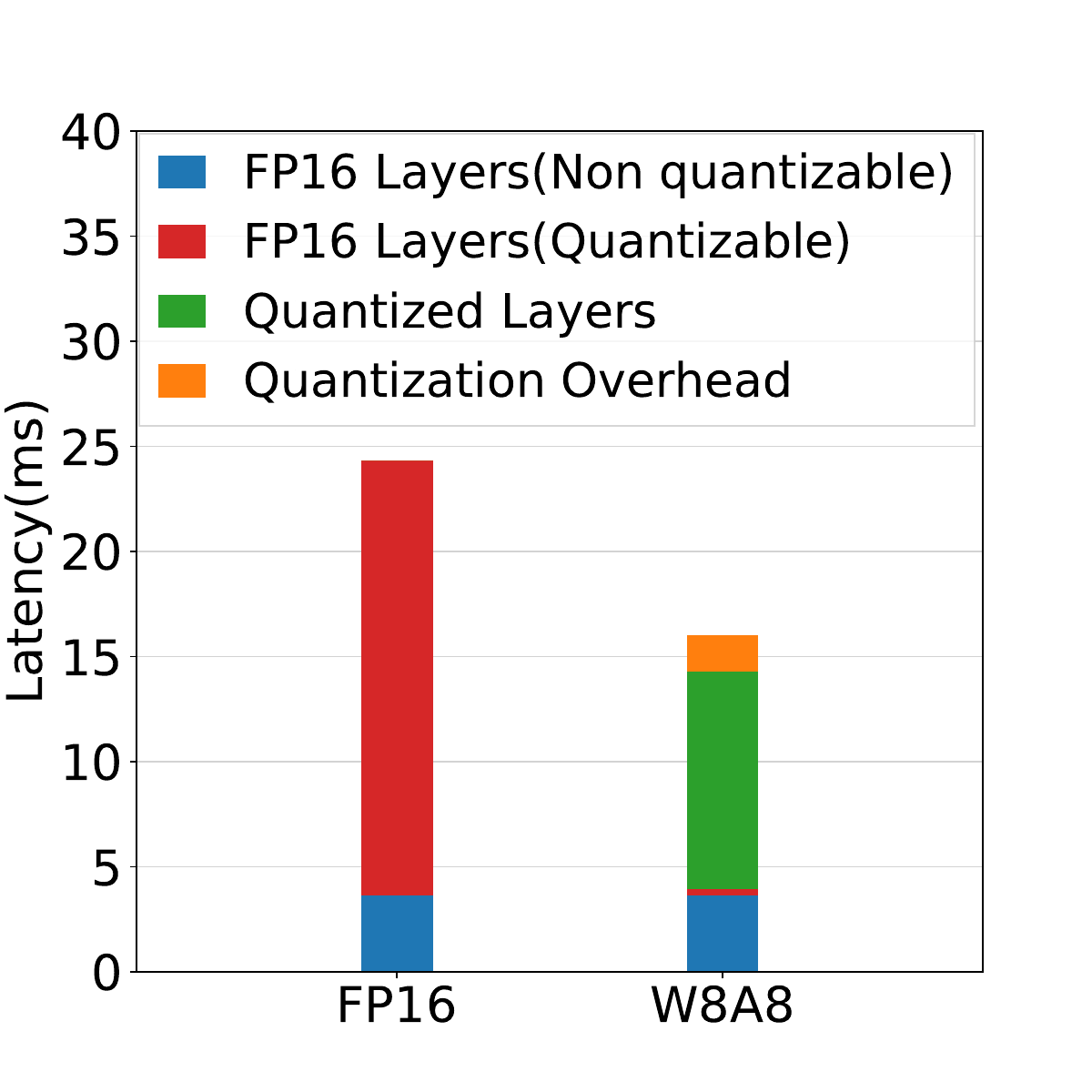}
            \caption{}
            \label{fig:hardware:b}
        \end{subfigure}%
    \end{minipage}
    \caption{\textbf{The illustration of MixDQ's hardware resource savings.} Left: The comparison of efficiency and performance under different MixDQ mixed precision configurations. Right: (a) MixDQ's optimization of peak memory, (b) MixDQ's latency breakdown under W8A8. }
    \vspace{-10pt}
    \label{fig:hardware}
\end{figure}

\textbf{Memory footprint reduction.} Fig.~\ref{fig:hardware} (a) shows the GPU memory usage of MixDQ and the FP16 baseline on UNet inference. MixDQ can reduce the memory footprint from two aspects. {Firstly}, quantizing the model weights leads to smaller allocated memory to store the UNet model. {Secondly}, quantizing the activations saves allocated memory to store residual connection activations. Combining the two benefits, we can effectively reduce the peak memory footprint by 1.87$\times$, 3.03$\times$ and 3.03$\times$ under the W8A8, W4A16 and W4A8, respectively. 

\noindent
\textbf{Speedup of model inference.} Fig.~\ref{fig:hardware} (a) shows the latency of the UNet model inference on Nvidia RTX 4080. We use the W8A8 scenario to showcase the speedup. MixDQ W8A8 can accelerate the inference of UNet by 1.52$\times$ over the FP16 baseline. \ref{fig:hardware} (b) gives a breakdown of the inference latency. There are three types of layers in the model: quantizable layers, including linear and convolution layers; non-quantizable layers, such as normalization and non-linear activation layers; and quantization layers, that perform the FP16-to-int8 conversion. The non-quantizable layers have the same latency in both the baseline and MixDQ. The quantizable layers are accelerated by 1.97$\times$, approximately the same ratio between INT8 and FP16 hardware peak throughput on RTX4080 (2$\times$). 
MixDQ requires quantization layers for converting activations, but even with this overhead, the end-to-end acceleration still reaches 1.52$\times$.





\begin{figure}
  \centering
  \begin{subfigure}[b]{0.30\textwidth}
    \centering
    \includegraphics[width=\textwidth]{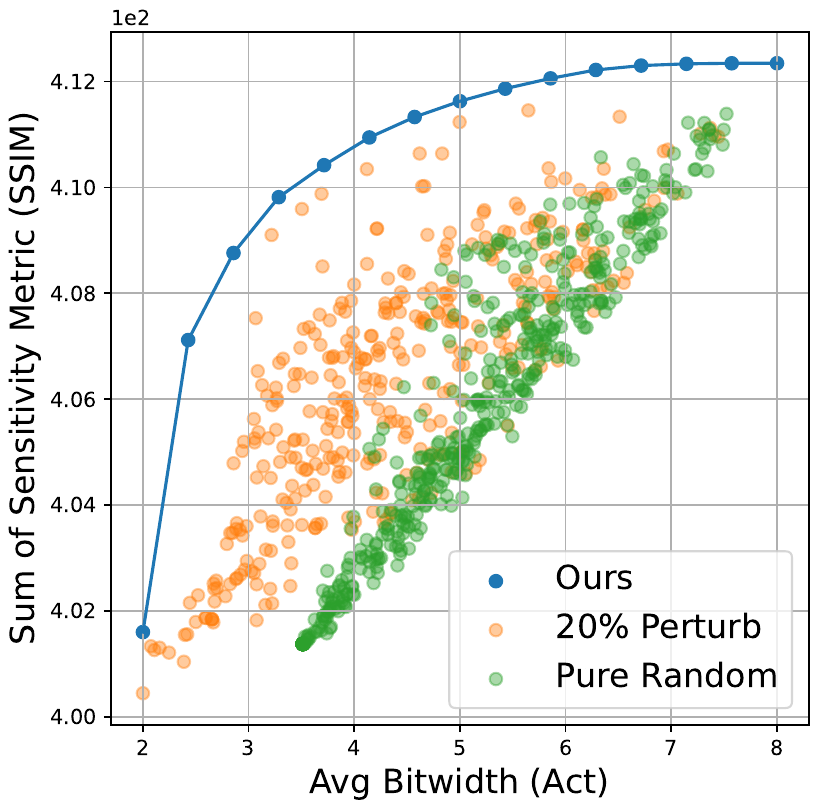}
    \caption{}
    \label{subfig:a}
  \end{subfigure}
  \hfill
  \begin{subfigure}[b]{0.30\textwidth}
    \centering
    \includegraphics[width=\textwidth]{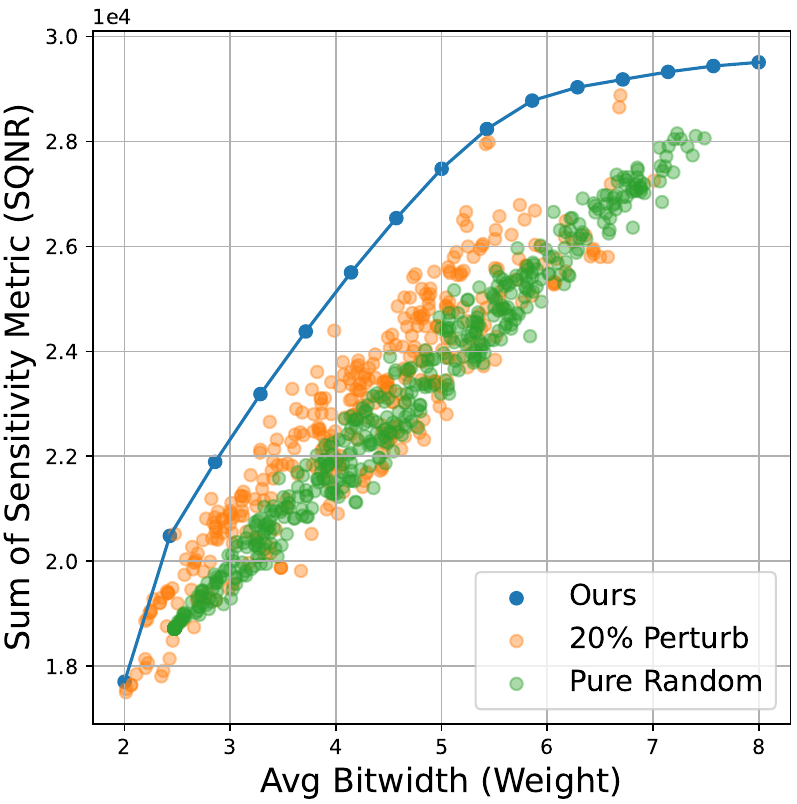}
    \caption{}
    \label{subfig:b}
  \end{subfigure}
  \hfill
  \begin{subfigure}[b]{0.30\textwidth}
    \centering
    \includegraphics[width=\textwidth]{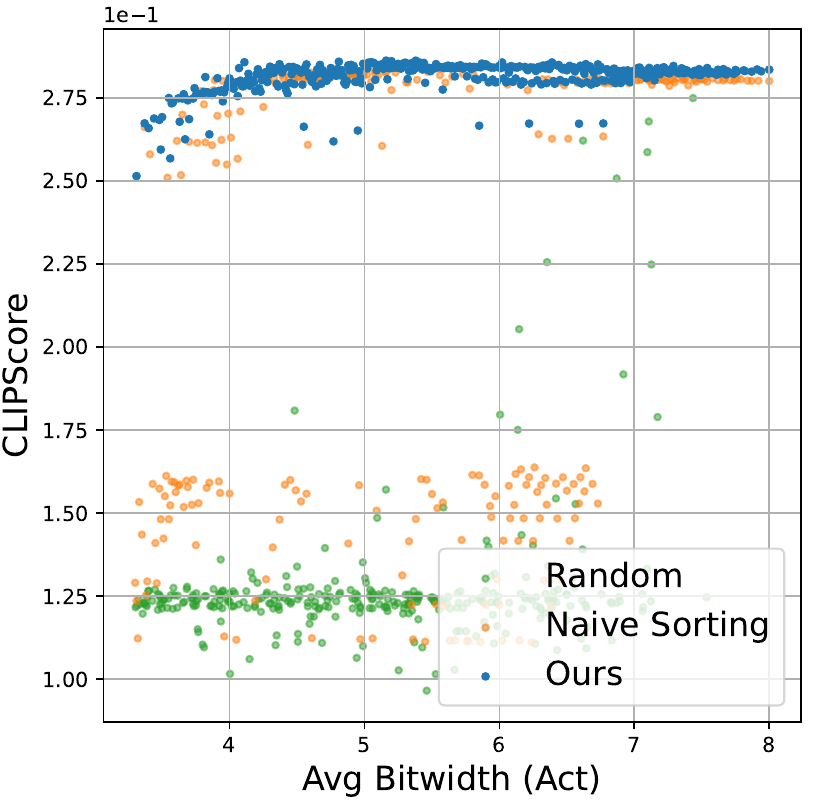}
    \caption{}
    \label{subfig:c}
  \end{subfigure}
    \caption{\textbf{The Pareto frontier of mixed precision configurations.} The x-axis represents the averaged bit-width, the y-axis of Subfigures (a), (b), (c) presents the sensitivity metrics (SQNR $\&$ SSIM), and the final evaluation metric (CLIPScore).
    }
    \vspace{-10pt}
    \label{fig:pareto}
\end{figure}

\section{Analysis}
\label{sec:analysis}

\subsection{Analysis of Paretor Frontier}
\label{sec:pareto}

The Pareto frontier~\cite{pareto-front} is often applied to present the accuracy-efficiency trade-off. We present the scatter plot of three metrics with respect to averaged bit-width for SDXL-turbo in Fig.~\ref{fig:pareto}. 
In Fig.~\ref{fig:pareto} (a),(b), we examine the effectiveness of integer planning by choosing the sensitivity metric (SQNR and SSIM) as the y-axis. As can be seen, compared with the ``Pure random'' baseline that randomly chooses bit-width, the ``Ours'' configuration achieves a significantly better trade-off. Furthermore, we randomly ``perturb'' 20\% of layers' bit-width in our configuration by replacing it with another bit-width and observing lower metric scores, denoting the superiority of our bit-width allocation.
Furthermore, to demonstrate that our integer programming is non-trivial, we design a ``Naive Sorting'' baseline that repeatedly lowers the bit-width of the least sensitive layer to fit the average bit-width. We compare the final evaluation metric (CLIP Score) of ``Ours'' result, ``Naive Sorting,'' and ``Random'' in Fig.~\ref{fig:pareto} (c). As can be seen, our result achieves a superior trade-off, demonstrating not only the effectiveness of the integer programming but also the accuracy of our acquired sensitivity.


\subsection{Analysis of BOS-aware Quantization}

\begin{figure}[h]
    \centering
    \includegraphics[width=0.8\textwidth]{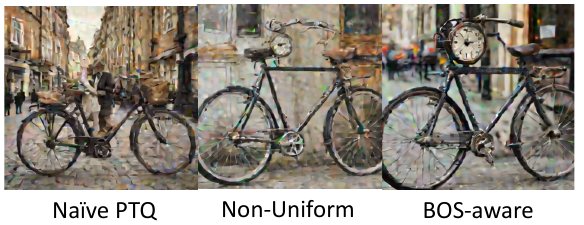}
    \caption{\textbf{The comparison of different quantization techniques for text embedding.} The content preservation notably improves for FP8 quantization, however, is still inferior to BOS-aware
    }
    \label{fig:bos_ablation}
\end{figure}

\begin{figure}[h]
    \centering
    \includegraphics[width=0.95\textwidth]{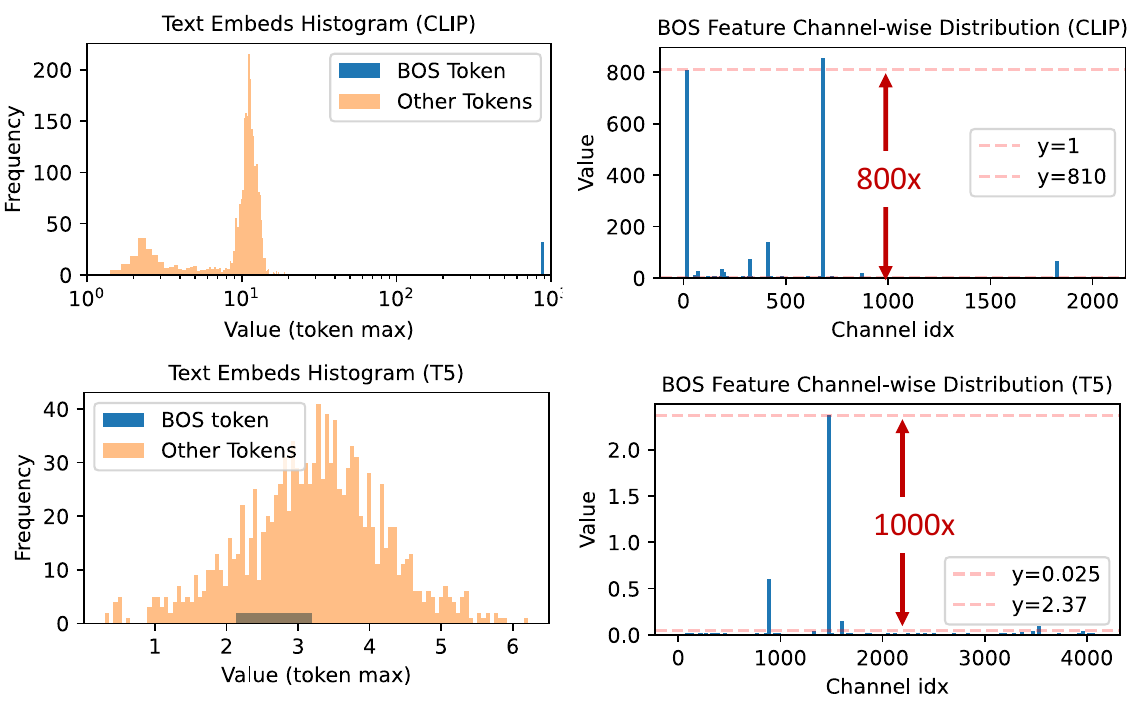}
    \caption{\textbf{The visualization of distribution for T5/CLIP text embeddings.} Left: the histogram for BOS and other tokens. Right: the channel-wise distribution of the BOS token.
    }
    \label{fig:bos_dist}
\end{figure}

We further conduct experiments to demonstrate the effectiveness of BOS-aware Quantization. As discussed in Section \ref{sec:bos}, the first BOS token in CLIP embedding exhibits significantly larger values compared to others. While some specialized quantization techniques, such as non-uniform quantization, could address this outlier issue, introducing FP8 quantization for text embedding still incurs notable performance degradation (See Fig. \ref{fig:bos_ablation}).
We also visualize the distribution of the BOS token and observe large variance along the channel dimension. Certain outlier channels have values 800 $\times$ greater than other channels, as depicted in Fig. \ref{fig:bos_dist}. Additionally, Fig. \ref{fig:bos_dist} illustrates that in the T5~\cite{t5} text encoder adopted by more recent papers \cite{pixart,para_to_image}, although the first token is no longer an outlier, the channel imbalance problem persists.
It's important to highlight that \textbf{BOS-aware quantization suggests that the quantization of the fixed BOS feature could be skipped}, resolving all its quantization issues beyond mere "outliers".

\subsection{Ablation Studies}
\label{sec:ablation}

\begin{figure}[h]
    \centering
    \includegraphics[width=0.9\textwidth]{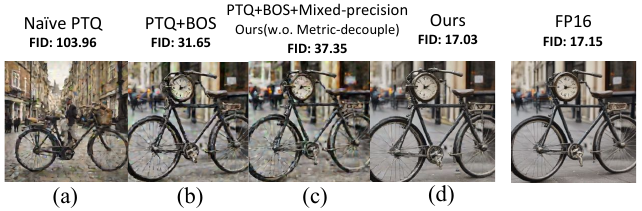}
    \caption{\textbf{The illustration of ablation studies on MixDQ techniques.} From left to right, as techniques are progressively added, notable improvement in both generated image quality and alignment is witnessed.}
    \vspace{-20pt}
    \label{fig:ablation}
\end{figure}

We conduct ablation studies by gradually incorporating MixDQ techniques into the W8A8 quantized SDXL-turbo model. As illustrated in Fig.~\ref{fig:ablation} and Tab.\ref{tab:ablation}, the generated images improve from severe degradation to surpassing FP ones. The Pareto frontier in Fig.~\ref{fig:pareto} could also assist in proving the effectiveness. We discuss the effectiveness of the proposed techniques individually as follows:

\noindent\textbf{Effectiveness of BOS-aware Quantization.} When adopting naive W8A8 PTQ for SDXL-turbo, as shown in Fig.~\ref{fig:ablation} and Tab.~\ref{tab:ablation}, the generated quality degrades, and the content changes significantly. After introducing BOS-aware quantization, the image content is recovered, and the metric values improve significantly (FID: 103.95 $\rightarrow$ 31.65, CLIP Score: 0.1478 $\rightarrow$ 0.2652).

\noindent\textbf{Effectiveness of Metric-Decoupled Sensitivity.} 
In Fig.~\ref{fig:ablation} (c), we present the result of the bit-width allocation based on SQNR without our metric-decoupled scheme. It can be observed that compared with Fig.~\ref{fig:ablation} (b), the quality worsened (FID: 31.65 $\rightarrow$ 37.35, CLIP Score: 0.2652 $\rightarrow$ 0.2624) after applying mixed precision. This reveals the insufficient accurate sensitivity and highlights the importance of our metric-decoupled technique. Also, Fig.~\ref{fig:pareto} (c)  helps prove the metric-decoupled sensitivity has a high correlation with final evaluation metrics.

\noindent\textbf{Effectiveness of Mixed Precision.} 
As shown in Tab.\ref{tab:ablation} and Fig.~\ref{fig:ablation} (d), when applying mixed precision with our metric-decoupled sensitivity analysis, MixDQ achieves lossless quantization with generated images nearly identical to FP and acquires similar metric values (FID: 17.03 vs. 17.15, CLIP Score: 0.2703 vs. 0.2722). Fig.~\ref{fig:pareto} (a),(b) also illustrates that our integer programming strikes the optimal performance-efficiency trade-off on the Pareto frontier.


\begin{table*}[h]
\centering
\caption{\textbf{Ablation studies on MixDQ techniques.} By gradually incorporating the proposed techniques on SDXL-turbo with W8A8 quantization, the generated images exhibit improvements from failure to surpassing the FP16 baseline.}\
\resizebox{0.8\linewidth}{!}{
\begin{tabular}{cccccc}
\toprule[1pt]
\textit{BOS-aware} & \quad \textit{Mixed-Precision} & \quad \textit{Metric-Decouple} \quad & Bit-width(W/A)  & FID ($\downarrow)$ & CLIP ($\uparrow)$ \\
\midrule \midrule
- & - & - & FP16 & 17.15 & 0.2722 \\
\midrule
- & - & - & 8/8 & 103.96 & 0.1478 \\
\checkmark & - & - & 8/8 & 31.65 & 0.2652 \\
\checkmark & \checkmark & - & 8/8 & 37.35 & 0.2624 \\
\checkmark & \checkmark & \checkmark & 8/8 & 17.03 & 0.2703 \\
\bottomrule[1pt]
\end{tabular}}
\label{tab:ablation}
\end{table*}

\subsection{Analysis of Quantization Method}

\noindent
\textbf{Takeaway Knowledge of Layer Sensitivity.} We conclude key takeaways of layer sensitivity for text-to-image diffusion models as follows: (1) The CrossAttn \& FFN affects image content while the SelfAttn \& Conv affect quality. This is illustrated in Fig.~\ref{fig:decouple} and Sec.~\ref{sec:metric-decouple}. (2) The activation for ``conv\_in'' and weight for ``conv\_out'' are highly sensitive compared with other convolution layers. (3) The sensitivity of activation for ``to\_k/v'' layers notably decreases after BOS-aware quantization, but their weight's sensitivity remain high. (4) In contrast to multi-step models, time embedding is not sensitive for few-step models. This fits intuition since distillation enables the network to denoise from any timestep. 

\noindent
\textbf{Properties of MixDQ Quantization.} (1) \textbf{Efficient Quantization}: Previous quantization methods require costly optimization of extensive AdaRound and scaling factor parameters, which takes tens of GPU hours. This process needs to be conducted repeatedly for different bit-width configurations.
MixDQ adopts the naive minmax quantization without bells and whistles. After obtaining sensitivity, it only requires several minutes to acquire the entire Pareto frontier. (2) \textbf{Hardware Friendly.} When designing quantization schemes, we prioritize making them hardware-friendly to minimize the hardware overhead for supporting them. This includes employing tensor-wise scaling for activations and utilizing output-channel-wise scaling for weights. The granularity of mixed precision is set at the layer level, enabling the utilization of existing kernels for hardware resource savings and avoiding the cost of implementing customized kernels. (3) \textbf{Flexibility.} The framework of MixDQ is flexible and extensible, allowing it to be combined with other quantization techniques such as efficient QAT~\cite{quest}.

\section{Conclusions}

We propose a mixed-precision quantization method: MixDQ, which consists of three steps. Firstly, BOS-aware text embedding quantization addresses the highly sensitive layers. Secondly, metric-decoupled sensitivity analysis is introduced to consider preserving both the image quality and content. Finally, an integer programming is conducted for bit-width allocation. MixDQ achieves W4A8 with negligible performance loss, and practical hardware speedup. 

\section*{Acknowledgements}

This work was supported by National Natural Science Foundation of China (No. 62325405, 62104128, U19B2019, U21B2031, 61832007, 62204164), Tsinghua EE Xilinx AI Research Fund, and Beijing National Research Center for Information Science and Technology (BNRist). We thank for all the support from Infinigence-AI.



\appendix
\newpage

\begin{center}
    \textbf{\Large Appendix}
\end{center}

\section{Application example of MixDQ}
\label{sec:Application example of MixDQ}

As discussed in the Sec. \ref{sec:integer-programming} and the Sec. \ref{sec:exp} in the main paper, we introduce mixed-precision bit-width allocation methods based on integer programming. This approach proves effective in obtaining points on the Pareto frontier. In this section, we further provide a detailed description of an application case: \textbf{``How to determine the optimal bit-width configuration given a certain memory budget''}?

We provide a detailed description for MixDQ in this application in \cref{algo:bitwidth-allocation}. For simplicity, the algorithm only describes the bit-width allocation for weights, a similar process could be applied for activation bit-width allocation. 
Firstly, we perform sensitivity analysis after BOS-aware quantization (discussed in the main paper Sec. \ref{sec:bos}). Following the idea of ``metric-decouple'', we categorize all layers of the model into content-related layers and quality-related layers. Specifically, the content-related layers include cross-attention layers and feed-forward networks (FFNs), while the quality-related layers encompass self-attention layers, convolutions, and other remaining layers (further discussed in \cref{sec:Justification of Metric Decouple}). 
Then we calculate the sensitivity separately for content/quality-related layers weight/activation, generating 4 groups of layer-wise relative sensitivities ($\mathcal{S}_{\mbox{content}}$ and $\mathcal{S}_{\mbox{quality}}$ in \cref{algo:bitwidth-allocation}). 





Subsequently, we allocate the bit-width under budget constraints based on the acquired sensitivity. Specifically, we describe the budget with averaged bit-width (e.g., W5A8) multiplied by the number of parameters for weight and activation, respectively (the $\mathcal{B_{\mbox{all}}}$). We linearly scan through $M$ nearby budgets $\mathcal{B}_j$ within range $[\mathcal{B}_{\mbox{all}} - \Delta \mathcal{B}, \mathcal{B_{\mbox{all}}}]$ to ensure we get the optimal result. Then, we introduce a hyperparameter $K$ (we choose linear space from 0.45 to 1,36 for weights, and 0.94 to 1.09 for activation) to describe the ratio between budgets assigned for the content-related and quality-related groups. It satisfies: $\mathcal{B}_{j,\mbox{all}}=\mathcal{B}_{\mbox{quality}}^{j,K} + \mathcal{B}_{\mbox{content}}^{j,K}$. After acquiring the budget for content or quality-related layers, integer programming could be used to get the mixed precision configuration: $\{b_{l,\mbox{content}}^{j,K}\}$, which is a layer-wise choice of candidate bit-width 2,4,8, the $l$ indexes the layers. The aforementioned process is iterated for each $B_{j,\mbox{model}}$ and $K$, and the optimal configurations are determined through a rapid evaluation of the generated images.

\begin{algorithm}
    \caption{\method Bit-width Allocation Algorithm}
    \label{algo:bitwidth-allocation}
    \resizebox{1.0\linewidth}{!}{ 
    \begin{minipage}{1\linewidth}
    \textbf{Input:} Sensitivity of content-related layers: $\mathcal{{S}}_{\text{content}}$,
                    Sensitivity of quality-related layers: $\mathcal{{S}}_{\text{quality}}$,
                    Parameter size of content-related layers : $\mathcal{P}_{\text{content}}$,
                    Parameter size of quality-related layers : $\mathcal{P}_{\text{quality}}$,
                    The target budget of the whole model, content and quality-related layers: $\mathcal{B}_{\text{all}}$, $\mathcal{B}_{\text{quality}}$,$\mathcal{B}_{\text{content}}$, 
                    The candidate budgets: $\{\mathcal{B}_{j}\} \in [\mathcal{B}_{\mbox{all}} - \Delta \mathcal{B}, \mathcal{B_{\mbox{all}}}]$ of length $M$, 
                    The candidate ratios: $\{{k}_i\}$ of length $N$, 
                    The FP model $\mathcal{M}$, the quantized model with bit-width configuration $\{b_{l}\}$: $\mathcal{M}_q^{\{b_{l}\}}$ 
    \textbf{Output:} The mixed precision configuration $\{b_l\}$.

    
    
    \For{$j = 1$ to $M$}{
        \For{$k = k_0$ to $k_{N-1}$}{

            \tcp{Split the budgets for content and quality-related layers.}
            
            $\mathcal{B}_{\text{content}}^{j, k}, \mathcal{B}_{\text{quality}}^{j, k} = SplitBudget(\mathcal{B}_{j,\text{all}}, \mathcal{P}_{content}, \mathcal{P}_{quality},k)$;

            \tcp{perform integer programming respectively}
            $\{{b}_{l, \text{content}}^{u, k}\} = IntegerProgramming(\mathcal{B}_{\text{content}}^{u, k}, S_{\text{content}}, \mathcal{P}_{\text{content}})$;
            
            $\{{b}_{l, \text{quality}}^{u, k}\} = IntegerProgramming(\mathcal{B}_{\text{quality}}^{u, k}, S_{\text{quality}}, \mathcal{P}_{\text{quality}})$;
            
            \tcp{get the mixed precision configuration for the whole model}
            $\{b_{l,\text{all}}^{u, k}\} = \{{b}^{u, k}_{l, quality},  {b}^{u, k}_{l, content}\}$;

            \tcp{infer with the quantized model}
            $ \text{imgs} = \mathcal{M_Q}^{\{b_{l,\text{all}}^{u, k}\}}(\text{prompts})$;

            \tcp{evaluate the generated images}
            $ \text{score}_{u,k} = eval(\text{imgs})$;
        }

            
  }  
    \tcp{choose the optimal configuration with best score}
    ${\{b_l\}} = \underset{\{b_{l,\text{all}}^{u, k}\}}{\mathrm{argmax}}{(\text{score}_{u, k})}$; 
  
  \Return $\{b_l\}$
  
  \end{minipage}
  }
\end{algorithm}

\section{Description of Challenges for Few-step Diffusion Quantization}

As mentioned in the main paper Sec. \ref{sec:intro} and Fig. \ref{fig:teaser}, Fig. \ref{fig:observation}, the few-step diffusion model quantization faces the challenges of (1) The few-step diffusion models are more sensitive to quantization than multi-step ones. (2) The quantization's effect on image content causes degradation in image-text alignment. In this section, we provide a detailed discussion of these two challenges. 

\subsection{Few-step Diffusion Models are More Sensitive to Quantization}

\begin{figure}[h]
    \centering
    \includegraphics[width=1\textwidth]{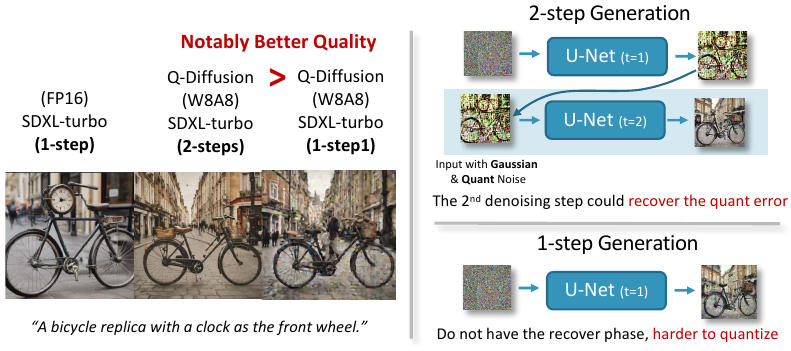}
    \caption{\textbf{The illustration of the few-step diffusion model is comparatively harder to quantize.} Left: the 2-step SDXL-turbo quantized model exhibits notably better image quality than the 1-step one. Right: the explanation of why the 1-step model is harder to quantize. 
    }
    \label{fig:challenge-few-step}
\end{figure}

As depicted in Fig. \ref{fig:teaser} of the main paper and discussed in Section 1, ``the 2-step model exhibits significantly less degradation compared to the SDXL-turbo 1-step model''. We delve into the reasons behind this phenomenon. Examining the challenge in Fig. \ref{fig:challenge-few-step}, we contrast the generation processes of the 2-step and 1-step models.

In the 2-step generation, the second step incorporates an image with both Gaussian noise and quantization noise as input. Recent literature on diffusion-based image editing, as highlighted in~\cite{sdedit,GLIDE}, indicates that diffusion models can effectively recover content from partially disrupted image input (image inpainting). Consequently, \textbf{the additional denoising steps (2nd or more) facilitate the reconstruction of quantization errors}, resulting in improved image quality.

In contrast, the 1-step diffusion model lacks this recovery phase, making it more ``vulnerable'' to quantization noise.

\subsection{Quantized Diffusion Models Lose Image-text Alignment}

\begin{figure}[h]
    \centering
    \includegraphics[width=1\textwidth]{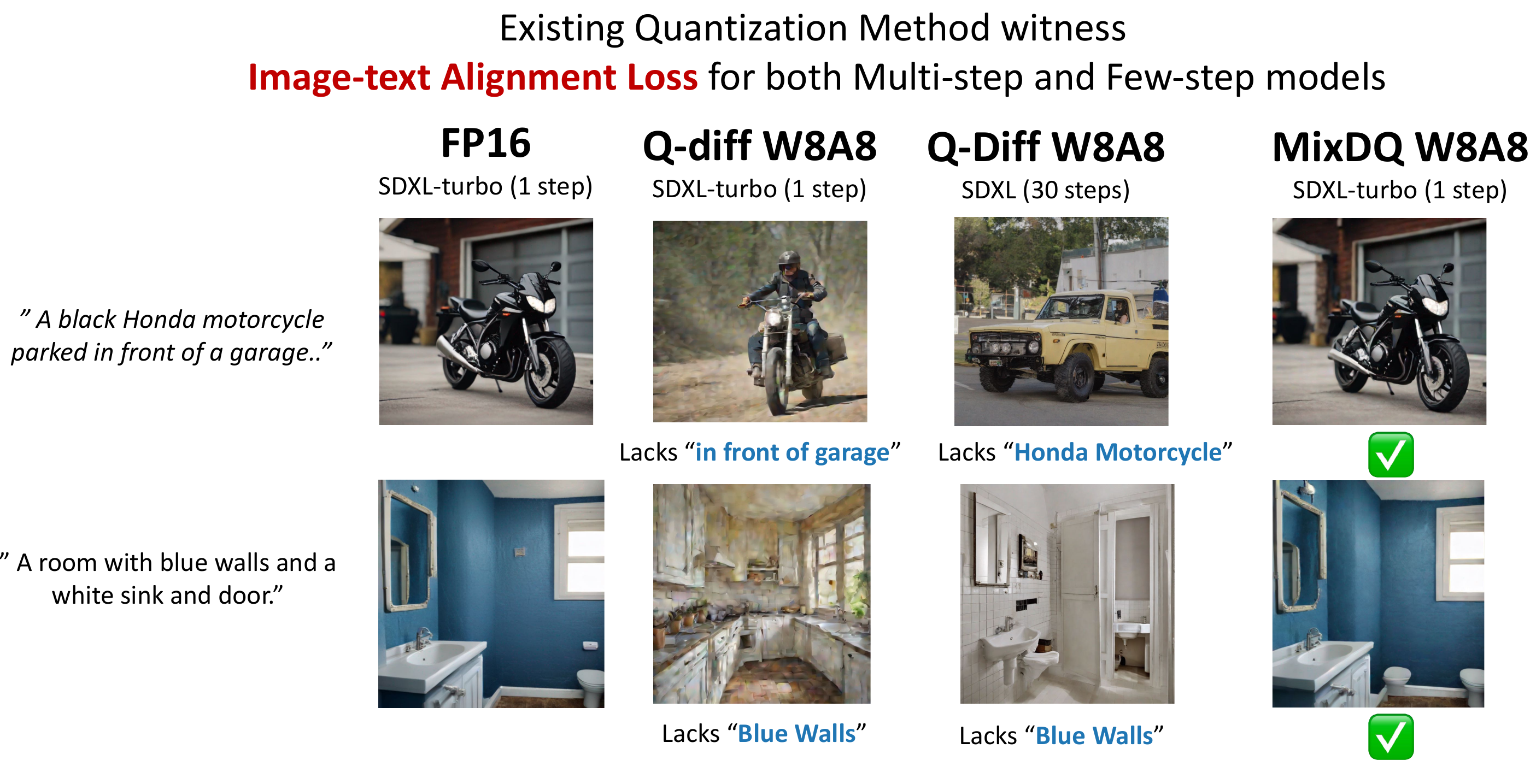}
    \caption{\textbf{The illustration of the ``Image-text Alignment loss'' challenge for diffusion quantization.} The Q-diffusion witnesses alignment loss (lacking of components described in the text instruction) for both the multi-step and few-step models. 
    }
    \label{fig:challenge-alignment}
\end{figure}

As outlined in Sec. \ref{sec:intro} of the main paper, prior research predominantly concentrates on preserving image quality. Nevertheless, it's crucial to recognize that quantization affects not just the image quality but also the content itself. The alteration in content can result in the degradation of image-text alignment. In Fig. \ref{fig:challenge-alignment}, we provide additional examples of alignment degradation, illustrating that such deterioration occurs not only in few-step models but also in less challenging multi-step diffusion models.

In the case of the prompt "A black Honda motorcycle parked in front of a garage," the W8A8 quantized SDXL-turbo 1-step model fails to include the "in front of garage" in the text instruction and instead generates an image of a man riding a bicycle. With the multi-step SDXL model, the output deviates even further, presenting a yellow truck instead of the specified black motorcycle.

Similarly, for the prompt "A room with blue walls and a white sink and door," both the W8A8 quantized models generate an image of a white room while omitting the "blue walls" component. In contrast, our MixDQ W8A8 produces images with \textbf{significantly improved image-text alignment}. They fulfill all components of the prompt, closely resembling the images generated by the FP16 model.

\section{Detailed Analysis of Hardware Experiments}
\label{sec:detailed_analysis_hardware_experiments}

In this section, we provide a detailed analysis of the hardware resource savings for different bit-width configurations and conclude our findings as follows:

(1) Illustrated in \cref{fig:hardware:b} in the main paper Sec. \ref{sec:exp:hardware}, \textbf{the W8A8 bar displays an almost negligible red segment}, corresponding to the "1\% most sensitive activation layers retained from FP16" as discussed in Sec. \ref{sec:exp:exp_performance} of the main paper. This component introduces minimal overhead while significantly contributing to the preservation of image quality.

(2) Observing the table in \cref{fig:hardware} in the main paper Sec. \ref{sec:exp:hardware}, we note that \textbf{W4A16 quantization solely achieves memory savings, while the latency remains consistent with the FP16 baseline}. The reduction in memory primarily stems from the decreased size of the model parameters, which are transferred to the GPU and serve as the "static memory cost." However, the W4A16 model continues to utilize FP16 computation, thus maintaining similar latency.

(3) Observing the table in \cref{fig:hardware} of the main paper Sec. \ref{sec:exp:hardware}, \textbf{W4A16 and W4A8 exhibit similar memory optimization}. The key distinction lies in the fact that W4A8 quantizes the activations, thereby reducing the memory cost through saved activations, primarily affecting the shortcut feature for U-Net. However, we discover that the occupied size by these activations is relatively small, amounting to less than 100MB compared to the model size of 6GB. Consequently, the additional savings attributed to activation quantization are not evident.

(4) Observing the table in \cref{fig:hardware} of the main paper Sec. \ref{sec:exp:hardware}, \textbf{both W4A8 and W8A8 demonstrate a noteworthy 1.52× latency speedup compared to the FP16 baseline}. This acceleration is attributed to the utilization of INT8 GPU kernels. It's worth noting that we haven't implemented the INT4 operators, making W4A8 and W8A8 quite similar in terms of latency improvement. However, W4A8 has the potential to achieve even higher latency speedup.

\section{Justification of Layer Grouping in Metric Decouple}
\label{sec:Justification of Metric Decouple}

As mentioned in the main paper Sec. \ref{sec:metric-decouple}, we split the layers into two groups based on their effect on image content and quality. Specifically, the cross-attention and feed-forward layers are regarded as the ``content-related layers'', and the convolution and self-attention layers as the ``quality-related layers''. In this section, we present experimental results to \textbf{demonstrate the rationale of the layer grouping}. 

\begin{figure}
    \centering
    \includegraphics[width=0.5\textwidth]{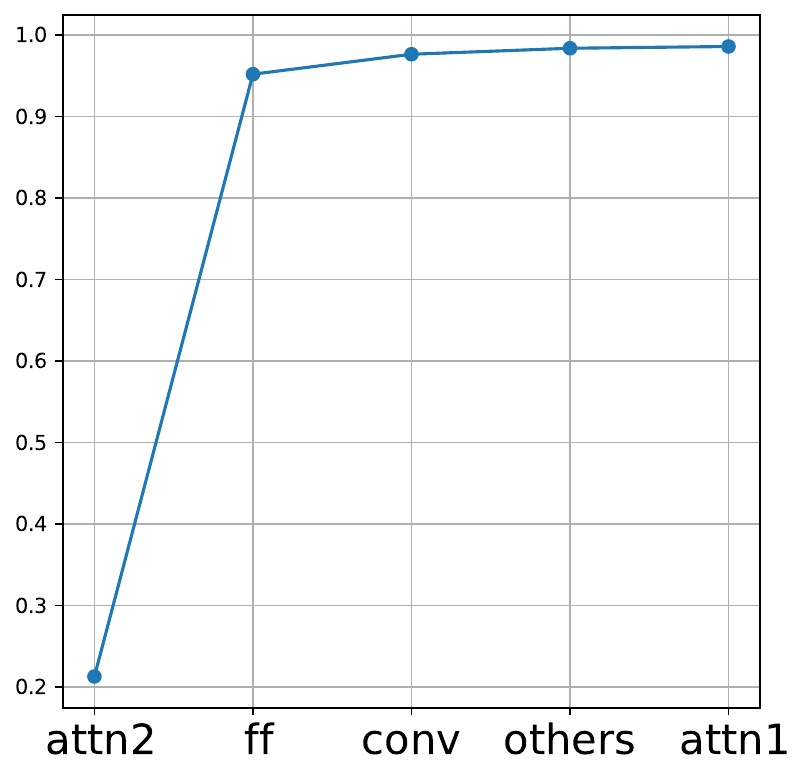}
    \caption{\textbf{SSIM Score when quantizing a certain group.} Cross-attention (CrossAttn) and Feed-Forward Networks (FFNs) are the two types of layers that have the greatest impact on content. 
    }
    \label{fig:group_ssim}
\end{figure}

Firstly, we summarize 5 layer types for the text-to-image diffusion model: cross-attention layers, feed-forward neural networks (FFNs), convolutions, self-attention layers, and other miscellaneous layers. We independently quantize the weight and activation into 8-bit for each layer type, and compute their effects on the SSIM score. As seen in \cref{fig:group_ssim}, the cross-attention and feed-forward layers have a significantly larger influence (0.8 and 0.05) on the image content compared with other layer types (less than 0.01). Therefore, we choose them as the ``content-related layers'', and the rest as ``quality-related'' layers.



\section{Qualitative Results}
\label{sec:Qualitative Result}

\begin{figure}[h]
    \centering
    \includegraphics[width=1\textwidth]{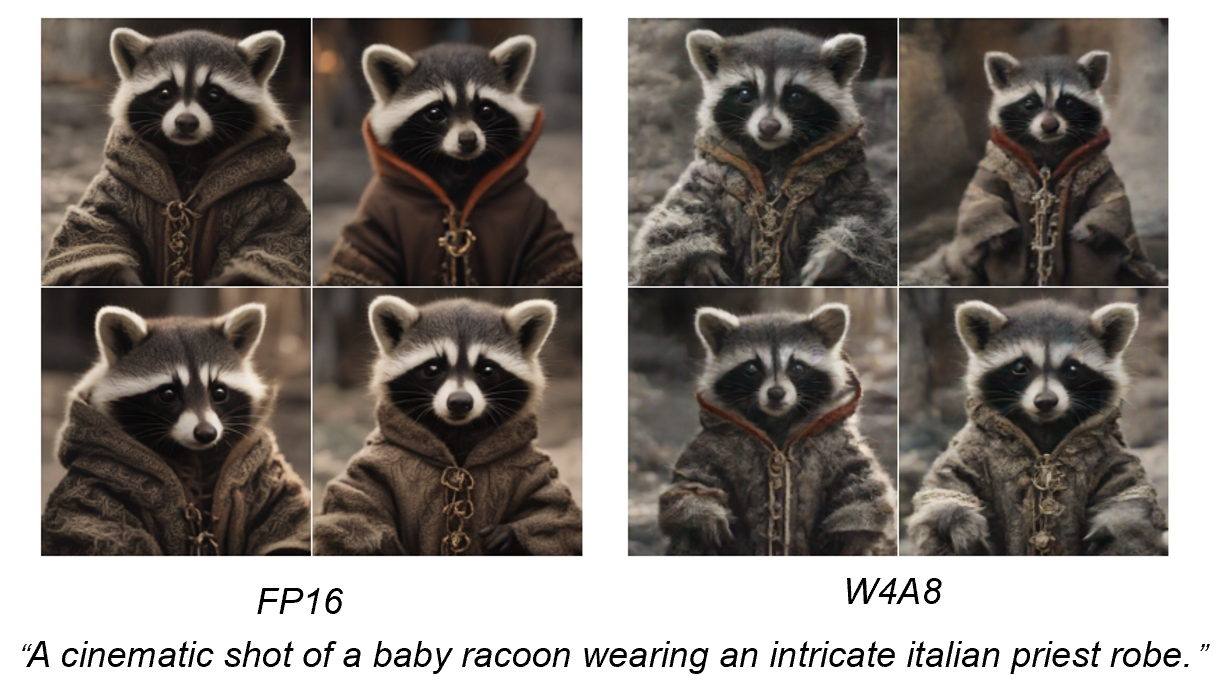}
    \caption{\textbf{The comparison between images generated by full precision model and quantized one.} Text-guided image generation from our W4A8 quantized SDXL-turbo with a fixed random seed. 
    }
    \label{cinematics}
\end{figure}

In this section, we provide more qualitative results to demonstrate the effectiveness of our MixDQ, especially in terms of \textbf{preserving the ``image-text alignment''}, 
As can be seen from \cref{cinematics}, even under W4A8, our method generates images with visual quality and image-text alignment similar to the FP16 model generated ones. 

\begin{figure}[h!]
    \centering
    \includegraphics[width=1\textwidth]{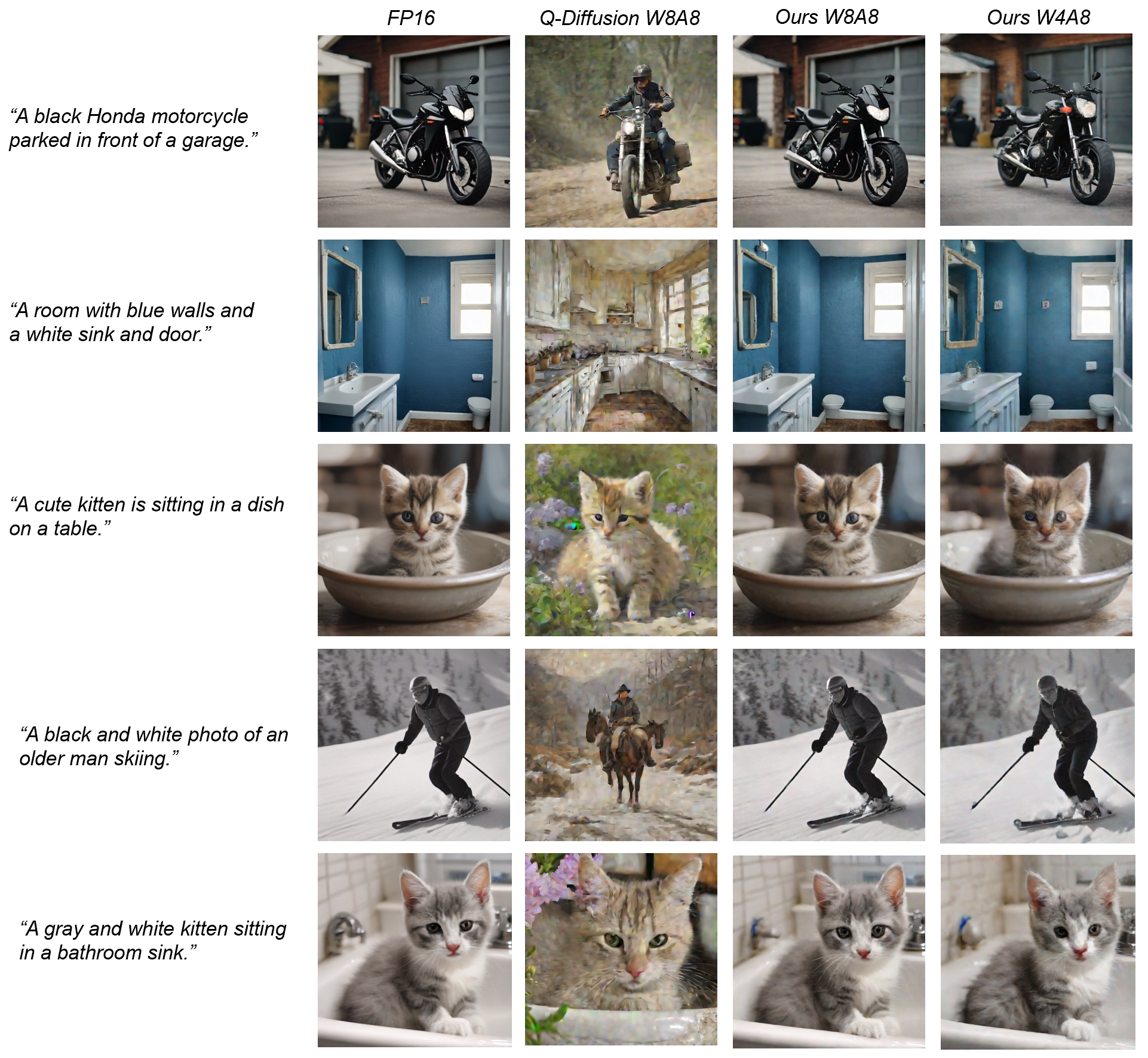}
    \caption{\textbf{The comparison between images generated by quatized model.} Our quantization scheme better maintains the content and quality of the images.
    }
    \label{qualitative results}
\end{figure}

\cref{qualitative results} highlights the superior text-image alignment achieved by MixDQ in comparison to baseline quantization methods. With W8A8 quantization, the Q-diffusion-generated images not only exhibit an "oil-painting" like blurring and tiny artifacts but also deviate from following the text instructions. For instance, it omits "in front of a garage" from the prompt and, instead, generates an image of a man riding a motorcycle. It also struggles to interpret ``an older man skiing''; instead, it generates a cowboy riding a strange two-headed horse. Moreover, the quantized model loses the capability to adhere to explicit color instructions like "blue walls" and "gray and white kitten." In contrast, with the more challenging W4A8 setting, our MixDQ quantized model produces images with nearly identical content compared to the FP16 baseline.


\hfill


\section{Limitations and Future Directions}

We introduce MixDQ, a mixed-precision quantization method that handles both the imbalance sensitivity and alignment degradation problems for diffusion quantization. We summarize the current limitations and potential future improvements as follows:

\begin{itemize}
    \item \textbf{Specialized Quantization Techniques for other Sensitive Layers.} In MixDQ, we pinpoint the bottleneck in text embedding quantization and craft BOS-aware quantization specifically tailored for it. Nevertheless, there are other highly sensitive layers, such as "conv-in" and "conv-out," that could also gain advantages from the design of specialized quantization techniques.
    \item \textbf{Combine with Improved Quantization Techniques.} In MixDQ, we employ naive min-max quantization as the quantization method. Its performance can be enhanced by incorporating advanced quantization techniques, such as Adaround~\cite{adaround}, or by introducing quantization-aware training~\cite{Jacob-quantization}.
    \item \textbf{Lower Bit-width.} In MixDQ, we opt for the 2, 4, and 8 as candidate bit-widths. When coupled with efficient lower-bit quantization methods, the mixed-precision bit-width allocation remains compatible with lower-bit-width combinations.
    \item \textbf{More Hardware Supports.} As discussed in \cref{sec:detailed_analysis_hardware_experiments}, we only utilize the INT8 GPU kernels for now, which restricts the performance for W4 quantization. The newest Nvidia TensorCore~\cite{tensorcore} supports INT4 computing, which could be further utilized for better efficiency. 
\end{itemize}




%
%
\clearpage
\bibliographystyle{splncs04}
\bibliography{main}
\end{document}